\documentclass[11pt,a4paper]{article}
\usepackage{times}
\usepackage{graphicx} 
\usepackage{subfigure}

\usepackage{natbib}

\usepackage{algorithm}
\usepackage{algorithmic}

\usepackage{booktabs} 
\usepackage{placeins} 
\usepackage[hyperref]{acl2017} 
\usepackage{xspace}
\usepackage{amsmath,amssymb,amsfonts,comment}
\usepackage{graphics,graphicx}
\usepackage{cleveref}
\usepackage{dsfont}
\usepackage{afterpage}
\usepackage{latexsym}
\usepackage{url}
\usepackage{multirow}

\aclfinalcopy 

\title{Learning when to skim and when to read}

\author{Alexander R. Johansen \and
        Richard Socher \\
        Salesforce Research, Palo Alto, CA, USA \\
        {\tt \{ajohansen,rsocher\}@salesforce.com}\\}
\date{}

\begin{document}
\maketitle
\begin{abstract}
Many recent advances in deep learning for natural language processing have come at increasing computational cost, but the power of these state-of-the-art models is not needed for every example in a dataset. We demonstrate two approaches to reducing unnecessary computation in cases where a fast but weak baseline classier and a stronger, slower model are both available. Applying an AUC-based metric to the task of sentiment classification, we find significant efficiency gains with both a probability-threshold method for reducing computational cost and one that uses a secondary decision network.
~\footnote{Blog post with interactive plots is available at \href{https://metamind.io/research/learning-when-to-skim-and-when-to-read}{https://metamind.io/research/learning-when-to-skim-and-when-to-read}}
\end{abstract}
\section{Introduction}
Deep learning models are getting bigger, better and more computationally expensive in the quest to match or exceed human performance~\cite{Wu2016,He2015,Amodei2015,Silver2016}.
With advances like the sparsely-gated mixture of experts~\cite{Shazeer2017}, pointer sentinel~\cite{Merity16}, or attention mechanisms~\cite{Bahdanau2015}, models for natural language processing are growing more complex in order to solve harder linguistic problems.
\begin{figure}[t]
\centering
\includegraphics[width=.99\linewidth]{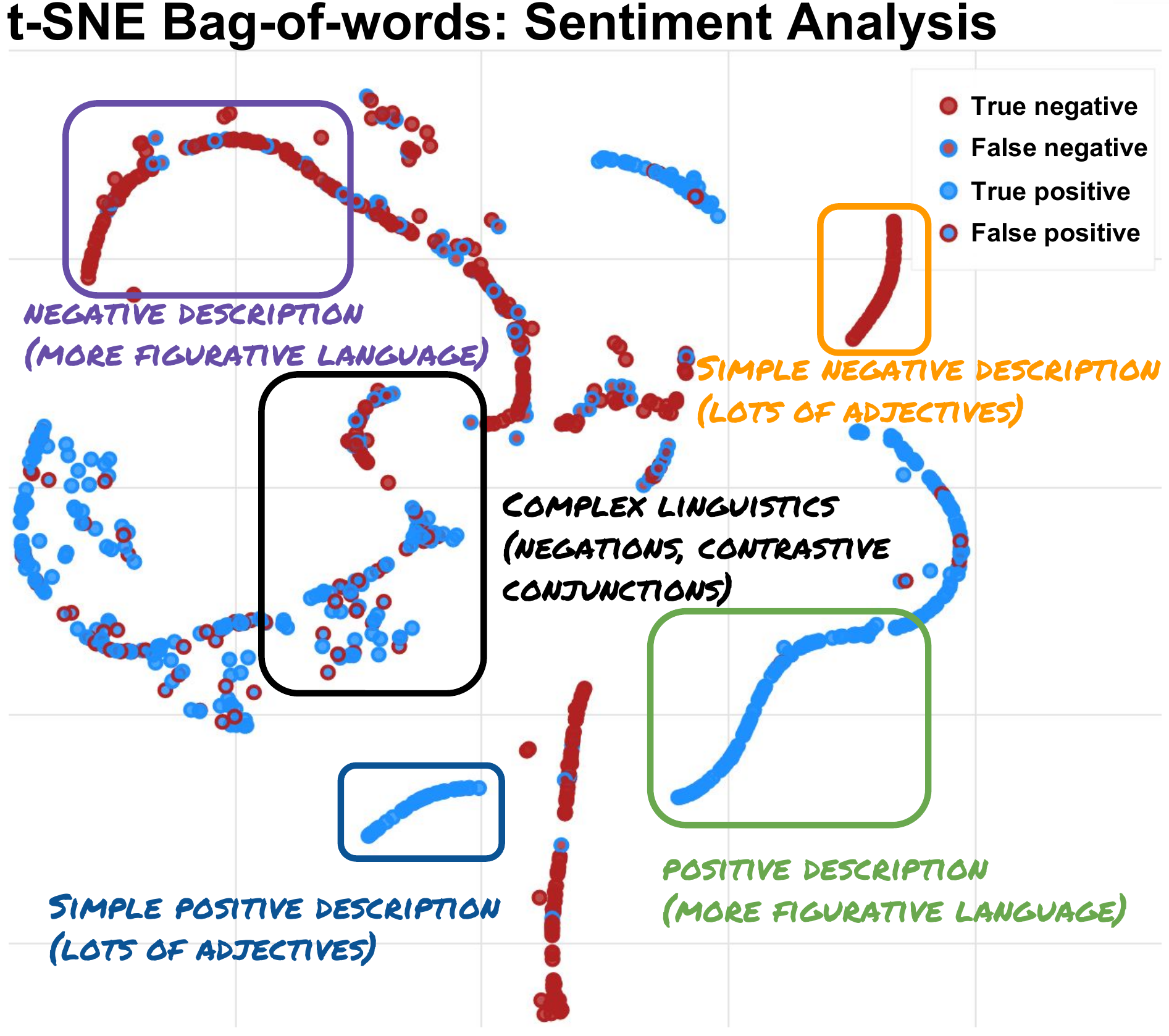}
\caption{Illustration with t-SNE~\cite{Van2008} of the activations of the last hidden layer in a computationally cheap bag-of-words (BoW) model on the binary Stanford Sentiment Treebank (SST) dataset ~\cite{Socher2013EMNLP}.
Each data point is one sentence, while the plot has been annotated with qualitative descriptions.}
\label{fig:frontimg}
\end{figure}
Many of the problems these new models are designed to solve appear infrequently in real-world datasets, yet the complex model architectures motivated by such problems are employed for every example.
For example, \cref{fig:frontimg} illustrates how a computationally cheap model (continuous bag-of-words) represents and clusters sentences.
Clusters with simple syntax and semantics (``simple linguistic content'') tend to be classified correctly more often than clusters with difficult linguistic content.
In particular, the BoW model is agnostic to word order and fails to accurately classify sentences with contrastive conjunctions.

This paper starts with the intuition that exclusively using a complex model leads to inefficient use of resources when dealing with more straightforward input examples.
To remedy this, we propose two strategies for reducing inefficiency based on learning to classify the difficulty of a sentence.
In both strategies, if we can determine that a sentence is easy, we use a computationally cheap bag-of-words (``skimming'').
If we cannot, we default to an LSTM (reading).
The first strategy uses the probability output of the BoW system as a confidence measure.
The second strategy employs a decision network to learn the relationship between the BoW and the LSTM.
Both strategies increase efficiency as measured by an area-under-the-curve (AUC) metric.
\section{When to skim: strategies}
To keep total computation time down, we investigate cheap strategies based on the BoW encoder.
Where the probability thresholding strategy is a cost-free byproduct of BoW classification and the decision network strategy adding a small additional network.
\subsection{Probability strategy}
\begin{figure}[t]
\centering
\includegraphics[width=.99\linewidth]{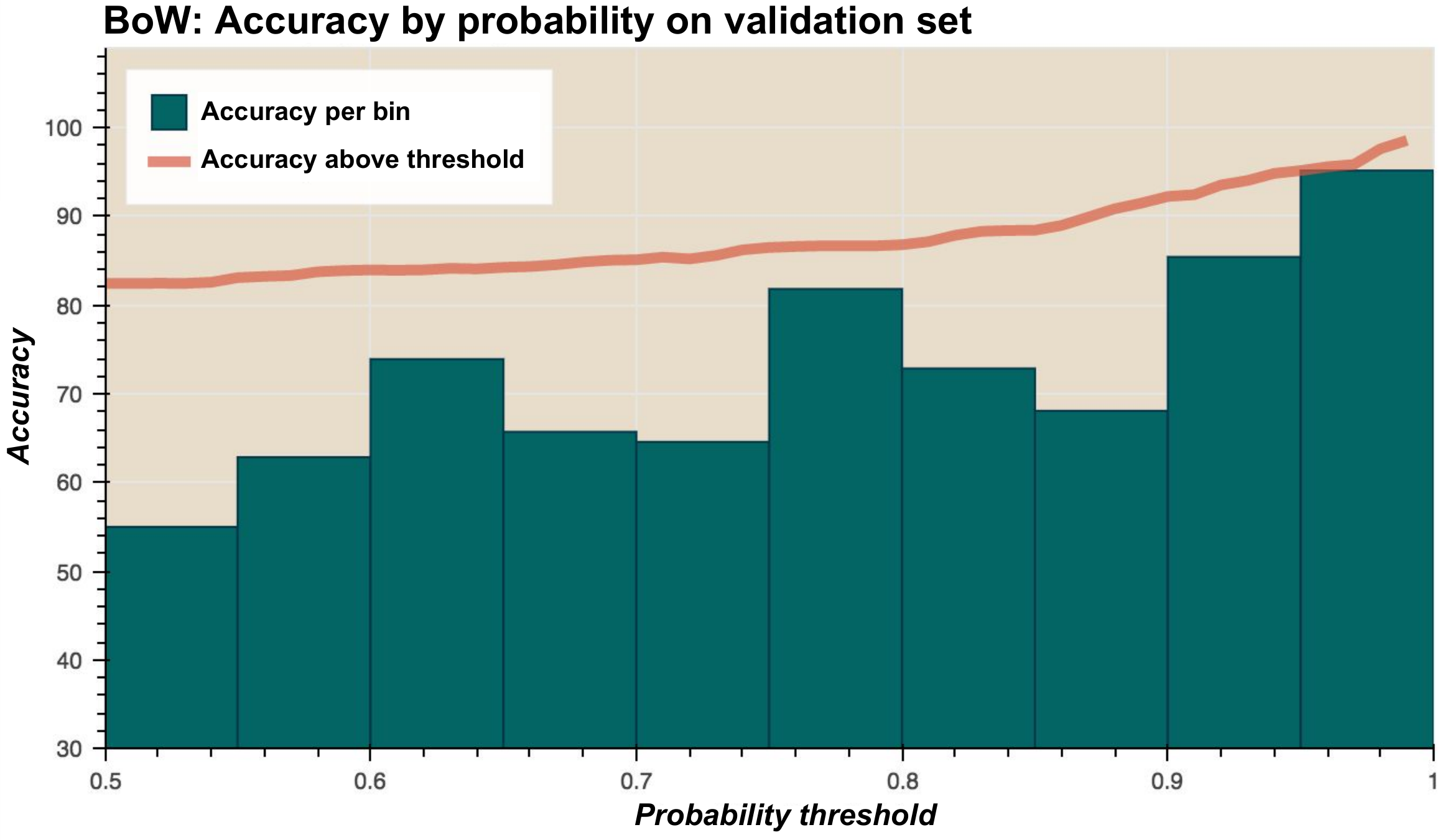}
\caption{Accuracy for thresholding on binary SST.
Probability thresholds are the maximum of the probability for the two classes.
The green bars corresponds to accuracy for each probability bucket (examples within a given probability span, e.g. $0.5<\Pr(Y|X, \theta_{\rm BoW})<0.55$), while the orange curve corresponds the accuracy of all examples with a probability above a given threshold (e.g. $\Pr(Y|X, \theta_{\rm BoW})<0.7$).
}
\label{fig:histo}
\end{figure}

\begin{figure}[t]
\centering
\includegraphics[width=.99\linewidth]{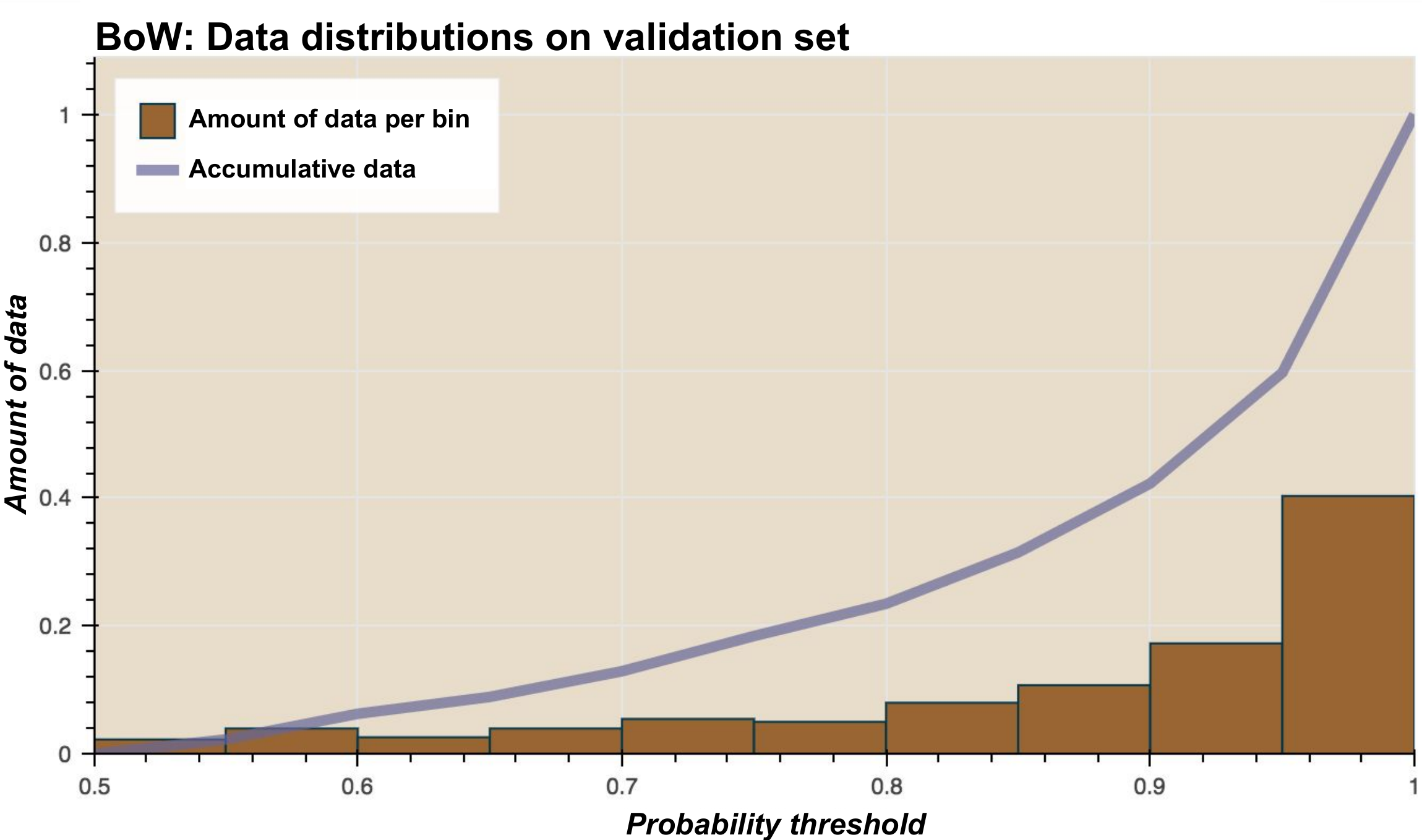}
\caption{Histogram showing frequency of each BoW probability bin on the SST validation set.
The line represents the cumulative frequency, or the fraction of data for which the expensive LSTM is triggered given a probability threshold.
}
\label{fig:dataamount}
\end{figure}

Since we use a softmax cross entropy loss function, our BoW model is penalized more for confident but wrong predictions. We should thus expect that confident answers are more likely correct.
\Cref{fig:histo} investigates the accuracy by thresholding probabilities empirically on the SST based on the BoW outputs, strengthening such hypothesis.
The probability strategy uses a threshold $\tau$ to determine which model to use, such that:
\begin{align*}
\hat{Y}_{\Pr(Y|X, \theta_{\rm BoW})>\tau} &= \arg\,\!\max_Y \Pr(Y|X, \theta_{\rm BoW})\\
\hat{Y}_{\Pr(Y|X, \theta_{\rm BoW})<\tau} &= \arg\,\!\max_Y \Pr(Y|X, \theta_{\rm LSTM})
\end{align*}
where $Y$ is the prediction, $X$ the input, $\theta_{\rm BoW}$ the BoW model and $\theta_{\rm LSTM}$ the LSTM model.
The LSTM is used only when the probability of the BoW is below the threshold.
\Cref{fig:dataamount} illustrates how much data is funneled to the LSTM when increasing the probability threshold, $\tau$.
\begin{table}[t]
\centering
\label{fig:conf}
\begin{tabular}{llcc}
\multicolumn{2}{c}{\multirow{2}{*}{\textbf{SST Valid}}} & \multicolumn{2}{c}{{BoW 82\%}} \\
\multicolumn{2}{l}{}                                & \multicolumn{1}{l}{\textbf{True}} & \textbf{False} \\
\multirow{2}{*}{{\begin{tabular}[c]{@{}l@{}}LSTM\\ 88\%\end{tabular}}} & \textbf{True}  & 76\%                              & 12\%           \\
         & \textbf{False} & 6\%       & 6\%
\end{tabular}
\caption{Confusion matrix for the BoW and the LSTM, where \textbf{True} means that the given model classifies correctly.}
\end{table}
\subsection{Decision network}
In the probability strategy we make our decision based on the expectation that the more powerful LSTM will do a better job when the bag-of-words system is in doubt, which is not necessarily the case.
\Cref{fig:conf} illustrates the confusion matrix between the BoW and the LSTM.
It turns out that the LSTM is only strictly better 12\% of the time, whereas 6\% of the sentences neither the BoW or the LSTM is correct.
In such case, there is no reason to run the LSTM and we might as well save time by only using the BoW.
\subsubsection{Learning to skim, the setup}
We propose a trainable decision network that is based on a secondary supervised classification task.
We use the confusion matrix between the BoW and the LSTM from \Cref{fig:conf} as labels.
We consider the case where the LSTM is correct and the BoW is wrong as the LSTM class and all other combinations as the BoW class.

\begin{figure}[t]
\centering
\includegraphics[width=.99\linewidth]{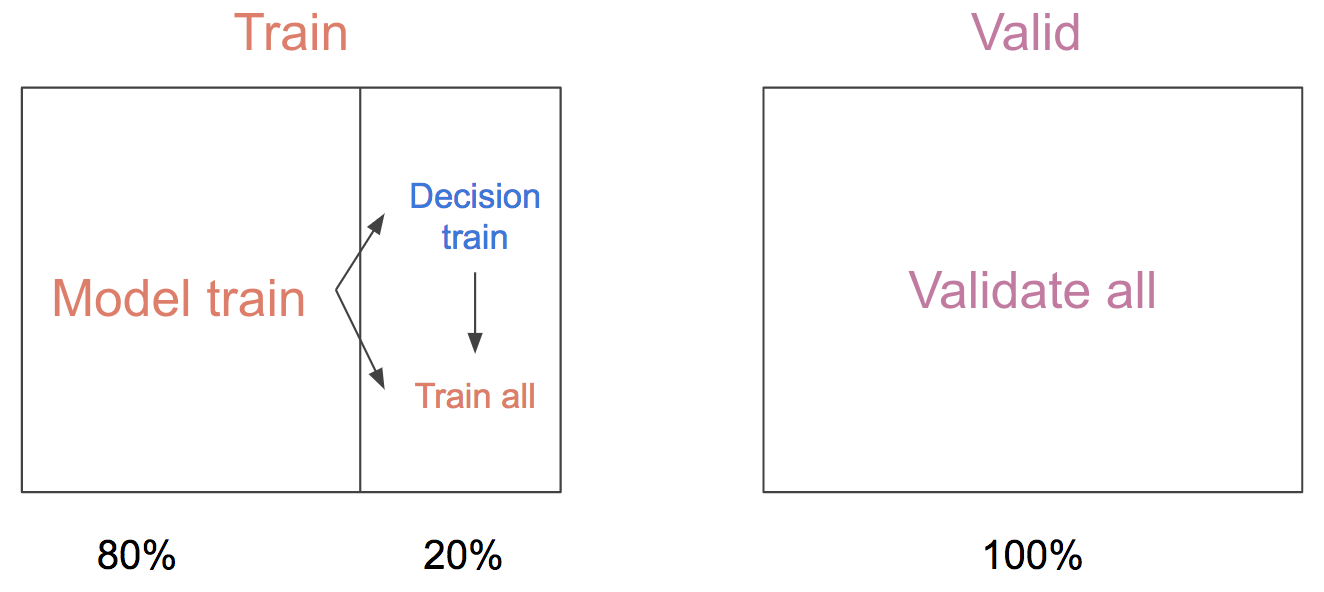}
\caption{We train both models (BoW and LSTM) on ``model train'', then generate labels and train the decision network on ``decision train'' and lastly fine tune the models on the full train set.
The full validation set is used for validation.
}
\label{fig:split}
\end{figure}

However, the confusion matrix on the train set is biased due to the models overfitting---which is why cannot co-train the decision network and our models (BoW, LSTM) on the same data.
Instead we create a new held-out split for training the decision network in a way that will generalize to the test set.
We split the training set into a model training set (80\% of training data) and a decision training set (remaining 20\% of training data).
We first train the BoW and the LSTM models on the model training set, generate labels for the decision training set and train the decision network on the decision training set, and lastly fine-tune the BoW and the LSTM on the original full training set while holding the decision network fixed.
We find that the decision network will still generalize to models that are fine-tuned on the full training set.
The entire pipeline is illustrated in \ref{fig:split}.
\section{Related Work}
\begin{table}[t]
\centering
\begin{tabular}{|l|l|}
\hline
\textbf{Model} & \textbf{Cost per sample} \\ \hline
Bag-of-words (BoW) & 0.16 ms                  \\
LSTM               & 1.36 ms                  \\ \hline
\end{tabular}
\caption{Computation time per sample for each model, with batch size 64.}
\label{tab:speed}
\end{table}
\begin{figure}[ht]
\centering
\includegraphics[width=.99\linewidth]{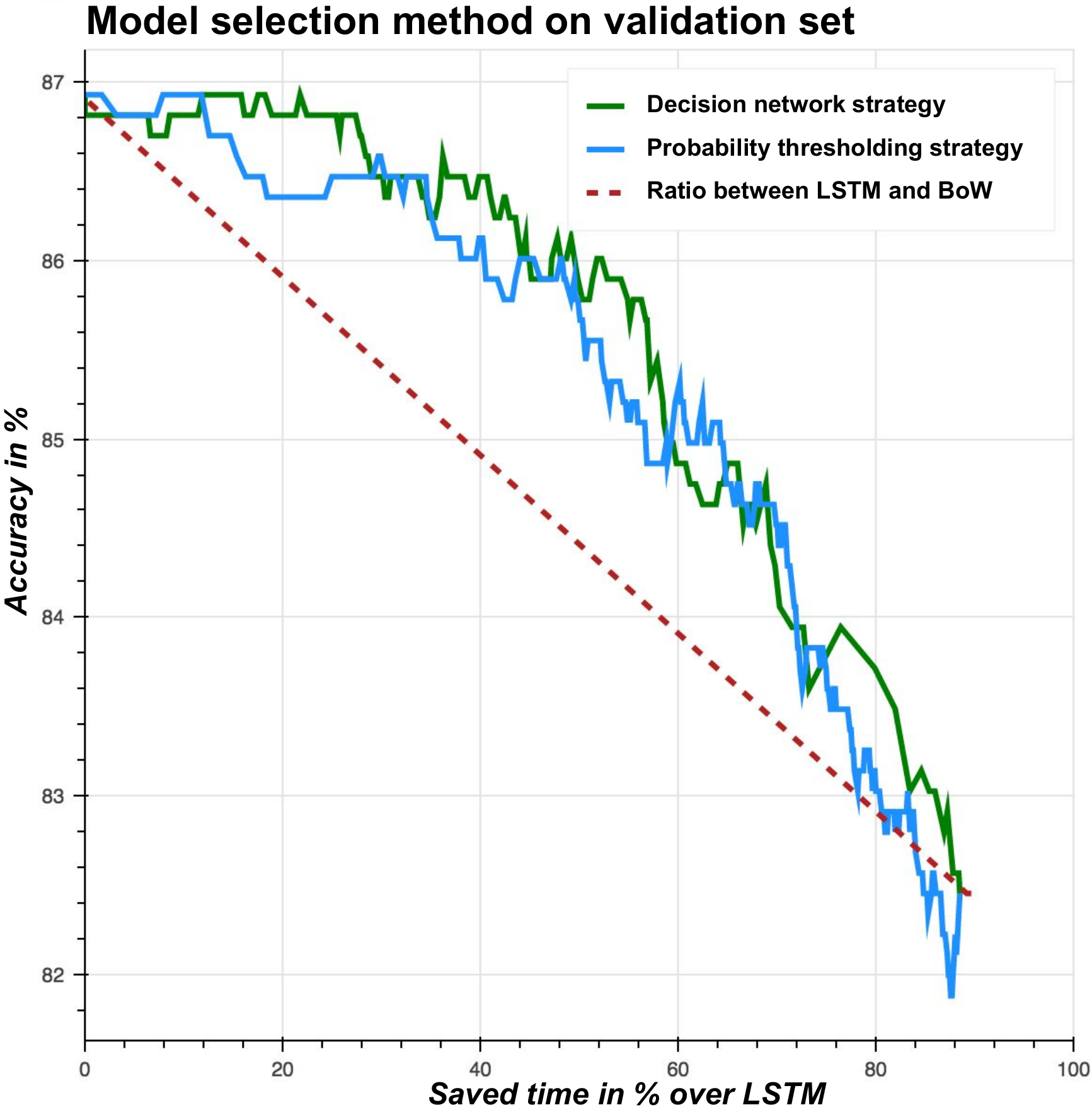}
\caption{Model usage strategies plotted with thresholds for the probability and decision network strategies chosen to save a given fraction of computation time.
The curve stops at around ~90\% savings as this represents using only the BoW model.
The dashed red line represents the na\"{i}ve approach of using the BoW and LSTM models at random with a fixed ratio.
}
\label{fig:dncurve}
\end{figure}
The idea of penalizing computational cost is not new.
Adaptive computation time (ACT)~\cite{Graves2016} employs a cost function to penalize additional computation and thereby complexity.
Concurrently with our work, two similar methods have been developed to choose between computationally cheap and expensive models.
\Citet{odena2017changing} propose the composer model, which chooses between computationally inexpensive and expensive layers.
To model the compute versus accuracy tradeoff they use a test-time modifiable loss function that resembles our probability strategy.
The composer model, similar to our decision network, is not restrained to the binary choice of two models.
Further, their model, similar to our decision network, does not have the drawbacks of probability thresholding, which requires every model of interest to be sequentially evaluated.
Instead, it can in theory support a multi-class setting; choosing from several networks
\Citet{BolukbasiWDS17} similarly use probability output to choose between increasingly expensive models.
They show results on ImageNet~\citep{imagenet_cvpr09} and provides encouraging time-savings with minimal drop in performance.
This further suggest that the probability thresholding strategy is a viable alternative to exclusively using SoTA models.
\begin{table*}[ht]
  \centering
  \begin{tabular}{|l|c|c|}
    \hline
    \textbf{Strategy} & \textbf{Validation AUC} & \textbf{Test AUC} \\ \hline
    Na\"{i}ve ratio & 84.84 & 83.77 \\
    Probability thresholding & $\mu=86.03, \sigma=0.3$ & $\mu=\textbf{85.49}, \sigma=0.3$ \\
    Decision network & $\mu=\textbf{86.13}, \sigma=0.3$ & $\mu=\textbf{85.49}, \sigma=0.3$ \\ \hline
    \end{tabular}
    \caption{Results for each decision strategy.
    The AUC is the mean value of the curve from \ref{fig:dncurve}.
    Each model is trained ten times with different initialization, and results are reported as mean and standard deviation over the ten runs.
    }
  \label{tab:suc}
\end{table*}
\section{Results}
\subsection{Model setup}
The architecture and training details of all models are all available in \cref{app:sup}.
In \cref{tab:speed} is an overview of the computational cost of our models.
Our dataset is the binary version of the Stanford Sentiment Treebank (SST), where ``very positive'' is combined with ``positive'', ``very negative'' is combined with ``negative'' and all ``neutral'' examples are removed.
\subsection{Benchmark model}
\label{ssec:benchmark}
To compare the two decision strategies we evaluate the trade-off between speed and accuracy, shown in \cref{fig:dncurve}.
Speedup is gained by using the BoW more frequently.
We vary the probability threshold in both strategies and compute the fraction of samples dispatched to each model to calculate average computation time.
We measure the average value of the speed-accuracy curve, a form of the area-under-the-curve (AUC) metric.

To construct a baseline we consider a na\"{i}ve ratio between the two models, i.e. let $Y_{\text{ratio}}$ be the random variable to represent the average accuracy on an unseen sample.
Then $Y_{\text{ratio}}$ has the following properties:
\begin{align}
\begin{cases}
\Pr(Y_{\text{ratio}}=A_\text{BoW}) = \alpha\\
\Pr(Y_{\text{ratio}}=A_\text{LSTM}) = 1-\alpha
\end{cases}
\end{align}
Where $A$ is the accuracy and $\alpha\in [0,1]$ is the proportion of data used for BoW.
According to the definition of the expectation of the random variable, we have 
the expected accuracy be:
\begin{align} \operatorname{E}(Y_{\text{ratio}}) &= \sum \Pr(Y_{\text{ratio}}=y) \times y\\
&= \alpha \times A_\text{BoW} + (1-\alpha) \times A_\text{LSTM} 
\end{align}
We calculate the cost of our strategy and benchmark ratio in the following manner.
\begin{align}
&C_{\text{strategy}} = C_{\text{BoW}} + (1-\alpha) \times C_{\text{LSTM}}\\
&C_{\text{ratio}} = \alpha \times C_{\text{BoW}} + (1-\alpha) \times C_{\text{LSTM}}
\end{align}
Where $C$ is the cost.
Notice that the decision network is not a byproduct of BoW classification and requires running a second MLP model, but for simplicity we consider the cost equivalent to the probability strategy.
\subsection{Quantitative results}
In \cref{fig:dncurve} and \Cref{tab:suc} we find that using either guided strategy outperforms the na\"{i}ve ratio benchmark by \textbf{1.72 AUC}.
\subsection{Qualitative results}
One might ask why the decision network is performing equivalently to the computationally simple probability thresholding technique.
In \ref{app:visualizations} we have provided illustrations for qualitative analysis of why that might be the case.
For example, \ref{fig:probdn} provides a t-SNE visualization of the last hidden layer in the BoW (used by both policies), from which we can assess that the probability strategy and the decision network follow similar predictive patterns.
There are a few samples where the probabilities assigned by both strategies differ significantly; it would be interesting to inspect whether or not these have been clustered differently in the extra neural layers of the decision network.
Towards that end, \ref{fig:compare} is a t-SNE plot of the last hidden layer of the decision network.
What we hope to see is that it learns to cluster when the LSTM is correct and the BoW is incorrect.
However, from the visualization it does not seem to learn the tendencies of the LSTM.
As we base our decision network on the last hidden state of the BoW, which is needed to reach a good solution, the decision network might not be able to discriminate where the BoW could not or it might have found the local minimum of imitating BoW probabilities too compelling.
Furthermore, learning the reasoning of the LSTM solely by observing its correctness on a slim dataset could be too weak of a signal.
Co-training the models in similar fashion to ~\citep{odena2017changing} might have yielded better results.
\section{Conclusion}
We have investigated if a cheap bag-of-words model can decide when samples, in binary sentiment classification, are easy or difficult.
We found that a guided strategy, based on a bag-of-words neural network, can make informed decisions on the difficulty of samples and when to run an expensive classifier.
This allow us to save computational time by only running complex classifiers on difficult sentences.
In our attempts to build a more general decision network, we found that it is difficult to use a weaker network to learn the behavior of a stronger one by just observing its correctness.
%
\bibliography{acl2017}
\bibliographystyle{acl_natbib}

\newpage
\appendix

\setcounter{figure}{0}
\renewcommand{\thefigure}{A\arabic{figure}}

\onecolumn
\section*{Supplementary material}
\label{app:sup}
\subsection*{Visualizations}
\label{app:visualizations}
t-SNE plots for the qualitative analysis section.
\begin{figure*}[ht]
\subfigure[Probability]{\includegraphics[width=.5\linewidth]{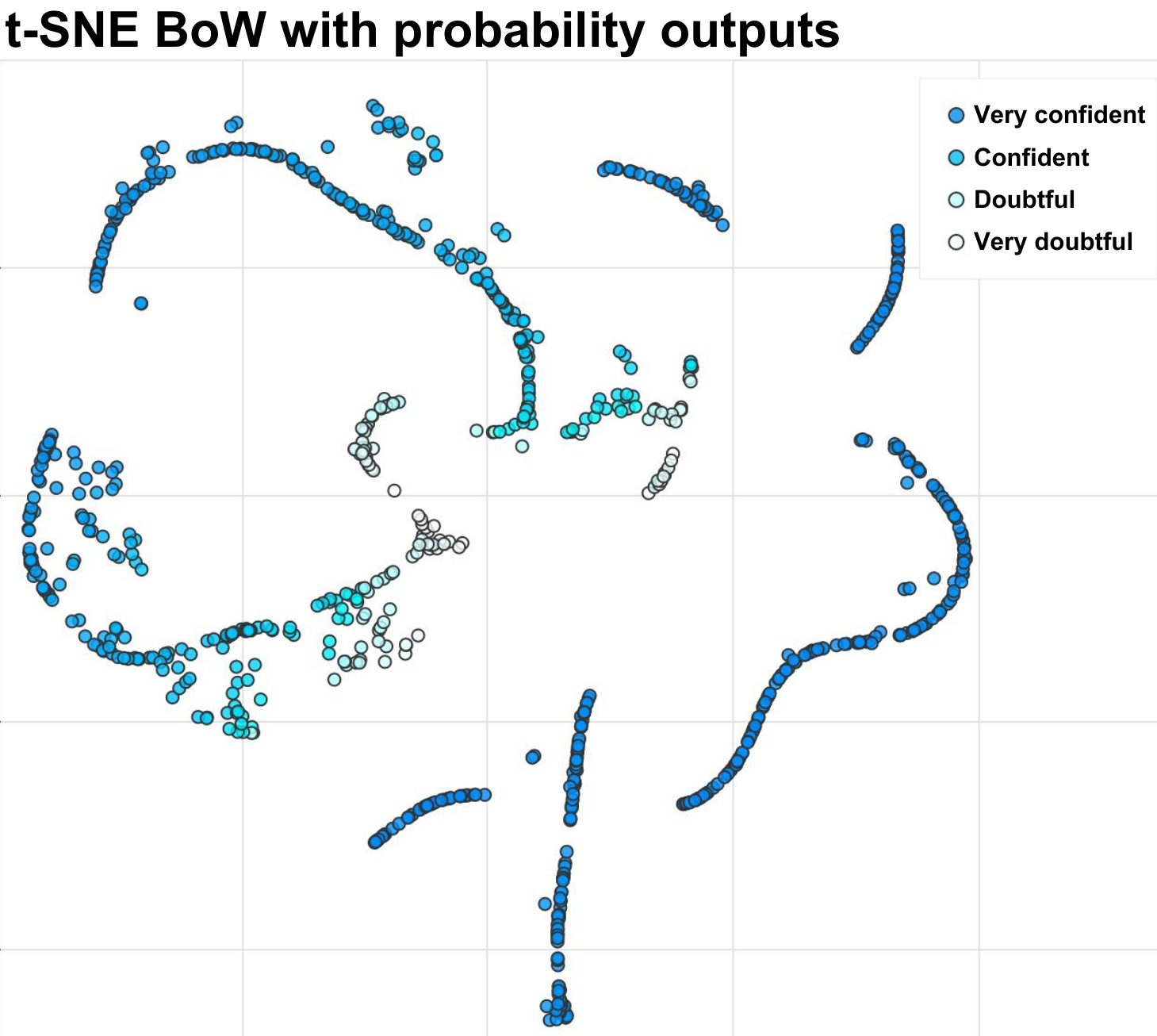}}
\subfigure[Decision Network]{\includegraphics[width=.5\linewidth]{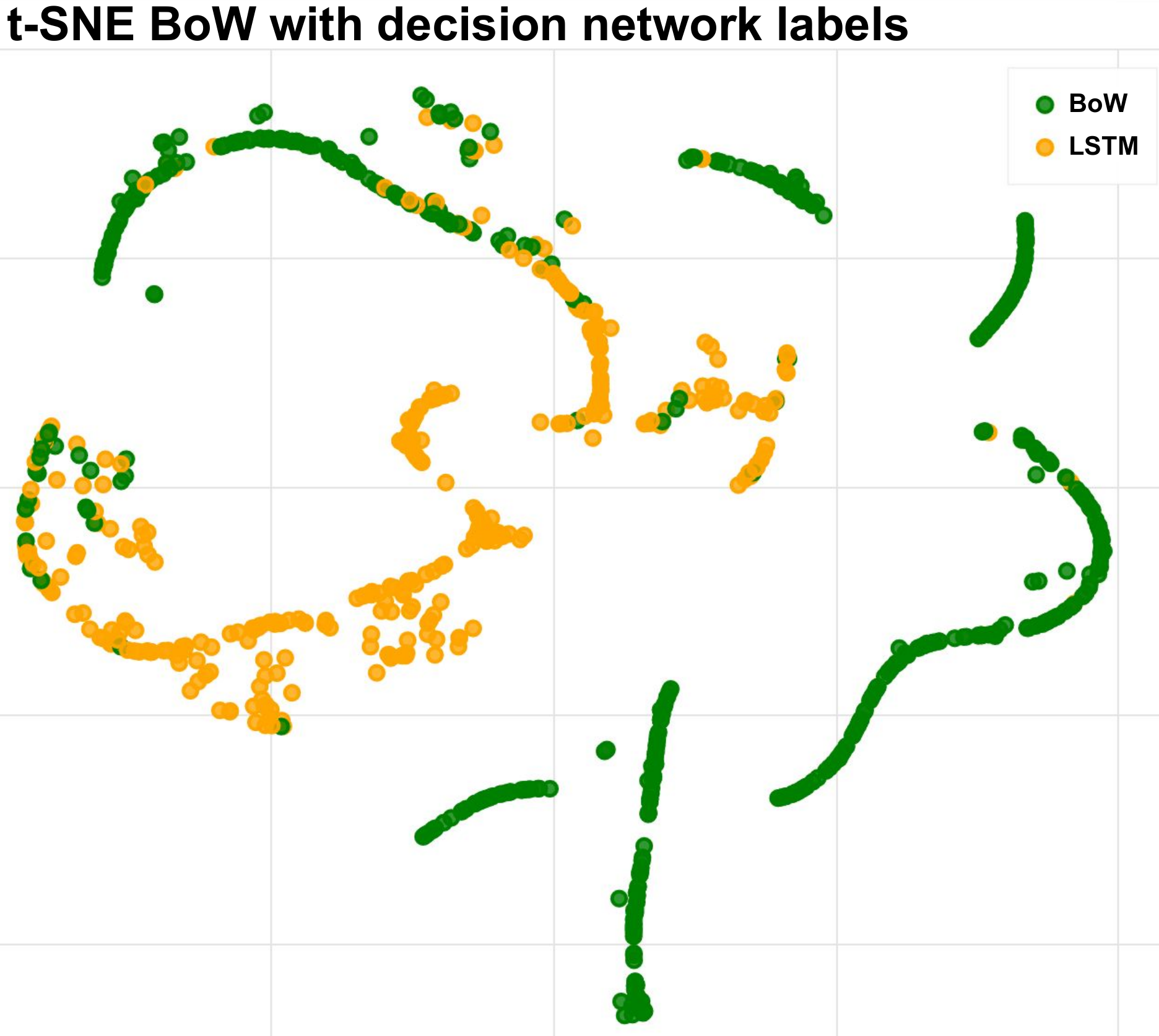}}
\caption{t-SNE plot of the last hidden layer of the BoW model.
The probabilities are colored by the confidence of the probability strategy.
The colors in the decision network plot are the predictions of which model should be used at a given threshold of the decision network.}
\label{fig:probdn}
\end{figure*}
\begin{figure}[ht]
\centering
\includegraphics[width=.99\linewidth]{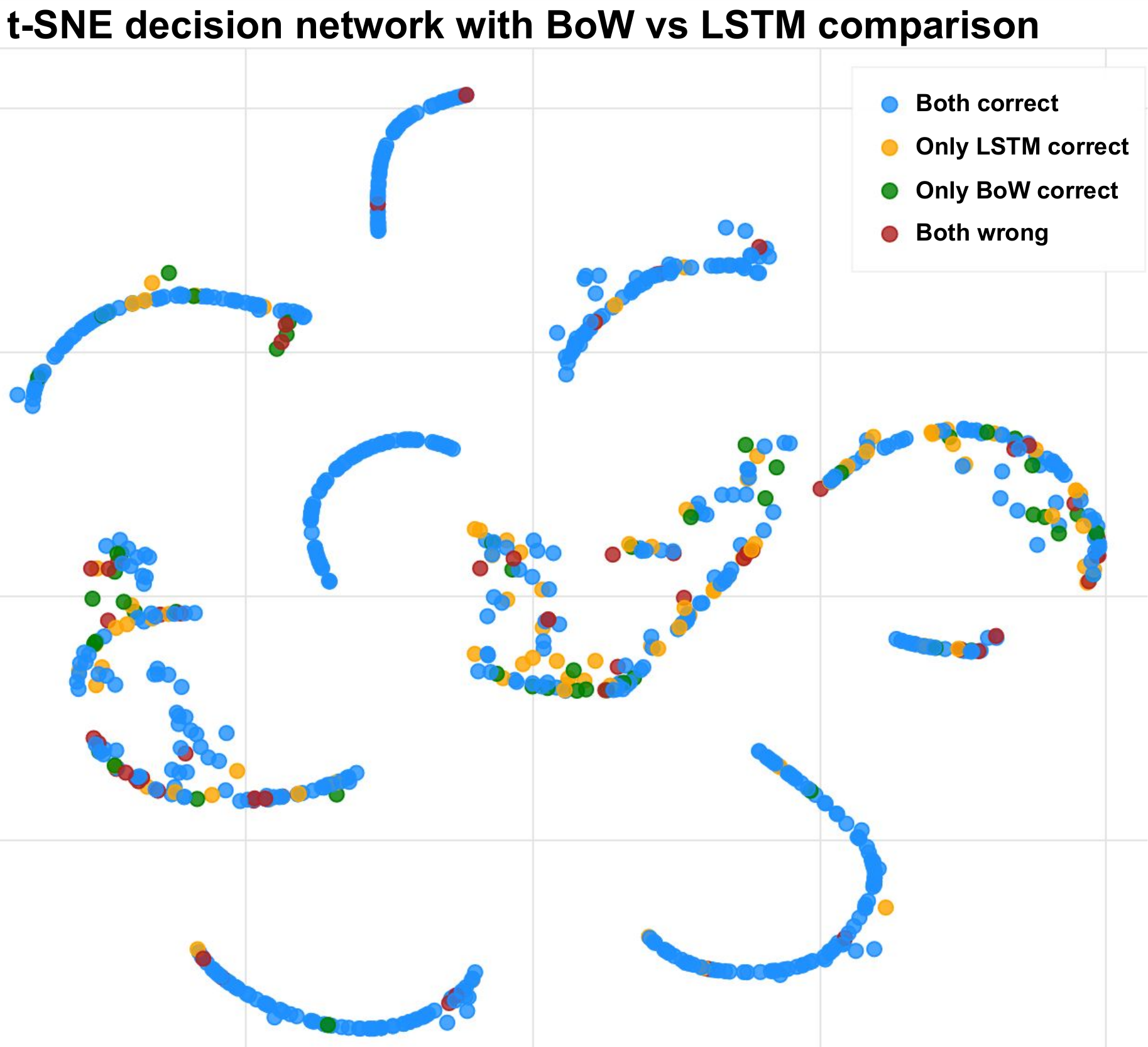}
\caption{We compare the predictions of the BoW and the LSTM to assess when one might be more correct than the other.
We train the decision network to separate the yellows (only LSTM correct) from the rest.
This plot enables us to evaluate if the model is able to learn the relationship between the correctness of the two models, even though it only has access to the BoW model.
}
\label{fig:compare}
\end{figure}
\subsection*{Model training}
\label{app:training}
All models are optimized with Adam~\cite{kingma2014adam} with a learning rate of $5\times 10^{-4}$.
We train our models with early stopping based on maximizing accuracy of all models, except the decision network where we maximize the AUC as described in \cref{ssec:benchmark}.
We use SST subtrees in both the model and decision train splits and for training both models.
\subsection*{Model illustration: The bag of words (BoW)}
\label{app:model}
\begin{figure}[t]
\hfill
\subfigure[LSTM]{\includegraphics[width=7.65cm]{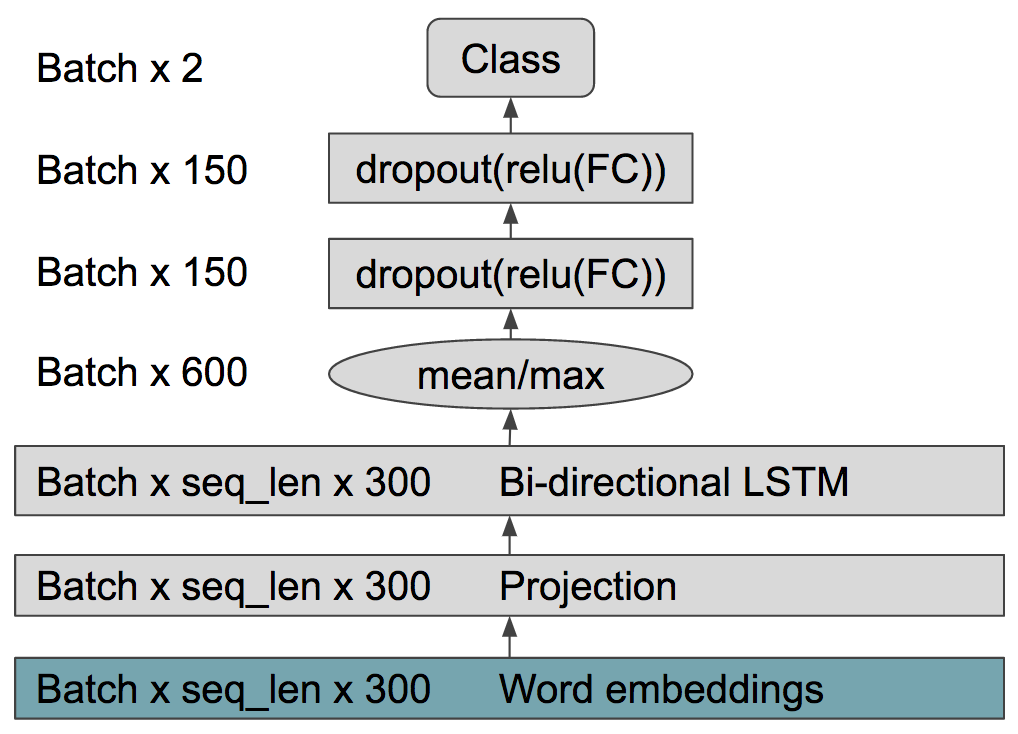}}
\hfill
\subfigure[Bag of words]{\includegraphics[width=8cm]{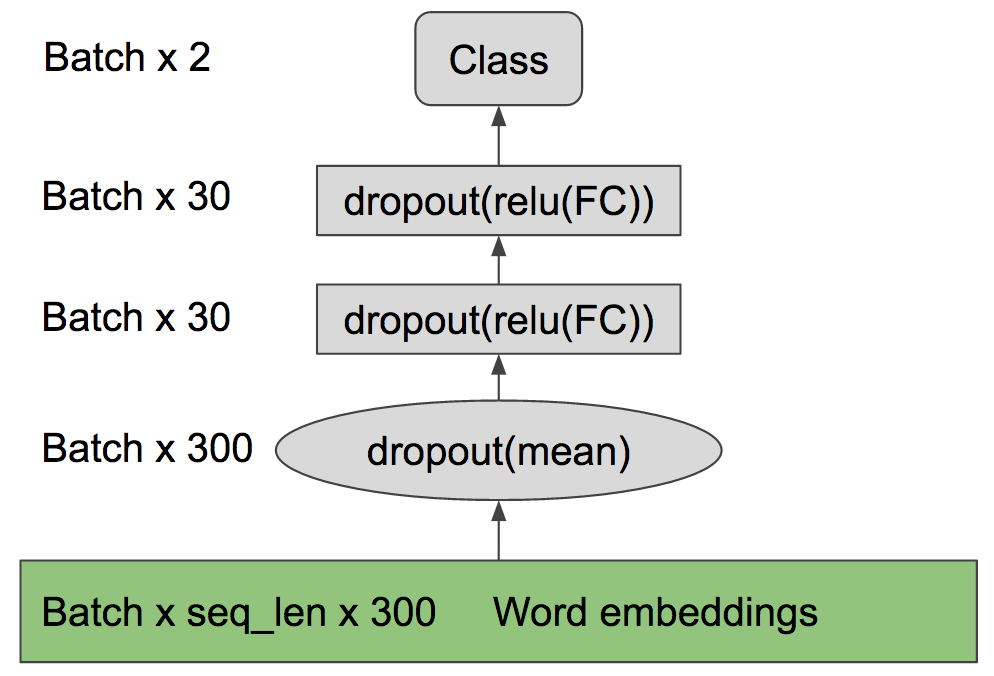}}
\hfill
\caption{Visualization of the BoW and LSTM model.
Green refers to initialization by GloVe and updating during training.
Grey is randomly initialized and updated during training.
Turquoise means fixed GloVe vectors.}
\label{fig:models}
\end{figure}

As shown in \cref{fig:models}, the BoW model's embeddings are initialized with pretrained GloVe~\cite{Pennington2014} vectors, then updated during training.
The embeddings are followed by an average-pooling layer~\cite{Mikolov2013b} and a two layer MLP~\cite{Ruck1990} with dropout of $p=0.5$~\cite{Srivastava2014}.
The network is first trained on the model train dataset (80\% of training data, as shown in ~\cref{fig:split}) until convergence (early stopping, at max 50 epochs) and afterwards on the full train dataset (100\% of training data) until convergence (early stopping, at max 50 epochs).
\subsection*{Model illustration: The LSTM}
The LSTM is visualized in \cref{fig:models}.
The LSTM's word embeddings are initialized with GloVe~\cite{Pennington2014}.
Instead of updating the embeddings, as is done in the BoW, we apply a trainable projection layer.
We find that this reduces overfitting.
After the projection layer a bi-directional~\cite{Schuster1997} recurrent neural network ~\cite{Graves2012} with long short-term memory cells~\cite{Hochreiter1997,Gers2000} is applied, followed by concatenated mean- and max-pooling of the hidden states across time.
We then employ a two layer MLP~\cite{Ruck1990} with dropout of $p=0.5$~\cite{Srivastava2014}.
The network is first trained on the model train dataset (80\% of training data) until convergence (early stopping, max 50 epochs) and afterwards on the full train dataset (100\% of training data) until convergence (early stopping, max 50 epochs).
\subsection*{Model illustration: The Decision Network}
\begin{figure}[t]
\centering
\includegraphics[width=.5\linewidth]{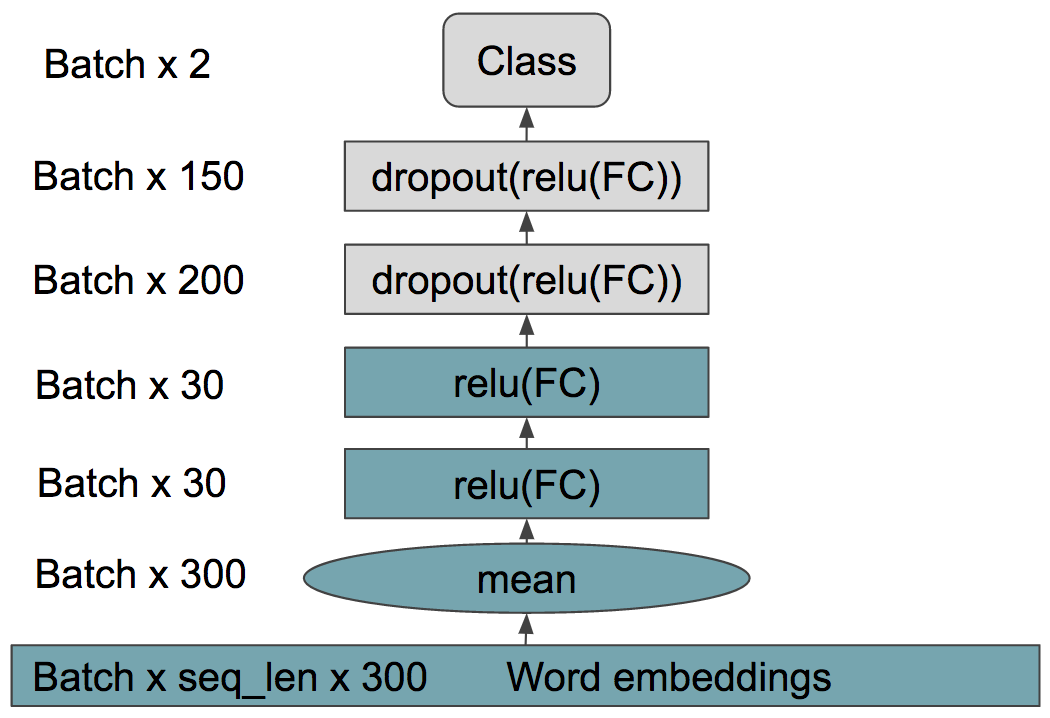}
\caption{Architecture of the decision network.
Turquoise boxes represent layers shared with the BoW model trained on the model train dataset and not updated during decision model training.}
\label{fig:decisionmodel}
\end{figure}

The decision network is pictured in \cref{fig:decisionmodel}, it inherits all but the output layer of the BoW model trained on the model train dataset, without dropouts.
The layers originating from the BoW are not updated during training.
We find that it overfits if we allow such.
From the last hidden layer of the BoW model, a two layer MLP~\cite{Ruck1990} with dropout of $p=0.5$~\cite{Srivastava2014} is applied on top.

The network is trained on the decision train portion of the dataset (20\% of training data) until convergence.
We use early stopping by measuring the AUC metric between the BoW and LSTM trained only on the model train dataset.
\end{document}